\documentclass[journal]{IEEEtran}
\makeatletter
\def\ps@IEEEtitlepagestyle{%
  \def\@oddfoot{\mycopyrightnotice}%
  \def\@oddhead{\hbox{}\@IEEEheaderstyle\leftmark\hfil\thepage}\relax
  \def\@evenhead{\@IEEEheaderstyle\thepage\hfil\leftmark\hbox{}}\relax
  \def\@evenfoot{}%
}
\def\mycopyrightnotice{%
  \begin{minipage}{\textwidth}
  \centering \scriptsize
  Copyright~\copyright~20XX IEEE.  Personal use of this material is permitted.  Permission from IEEE must be obtained for all other uses, in any current or future media, including reprinting/republishing this material for advertising or promotional purposes, creating new collective works, for resale or redistribution to servers or lists, or reuse of any copyrighted component of this work in other works.
  \end{minipage}
}
\makeatother

\ifCLASSINFOpdf
\else
\fi
\usepackage{amsmath}
\usepackage{multirow} 
\usepackage{multicol}
\usepackage{arydshln}
\usepackage{subfigure}
\usepackage{graphicx}
\usepackage{url}
\usepackage[numbers]{natbib}
\usepackage{amsmath}
\usepackage{amsfonts}
\usepackage{amssymb}
\usepackage{ragged2e}
\usepackage{subfigure}
\usepackage{pgfplots}
\usepackage{booktabs}
\usepackage{longtable}
\usepackage{pifont}
\newcommand{\cmark}{\ding{51}}%
\newcommand{\xmark}{\ding{55}}%
\begin{document}
%
\title{Channel-aware Decoupling Network for Multi-turn Dialogue Comprehension}
%
%
%

\author{Zhuosheng Zhang, Hai Zhao, Longxiang Liu
\thanks{Z. Zhang and H. Zhao are with the Department of Computer Science and Engineering, Shanghai Jiao Tong University, and also with Key Laboratory of Shanghai Education Commission for Intelligent Interaction and Cognitive Engineering, Shanghai Jiao Tong University, and also with MoE Key Lab of Artificial Intelligence, AI Institute, Shanghai Jiao Tong University. L. Liu is with Key Laboratory of Intelligent Information Processing
Institute of Computing Technology, Chinese Academy of Sciences (ICT/CAS). This work was conducted when L. Liu was with the Department of Computer Science and Engineering, Shanghai Jiao Tong University. E-mail: zhangzs@sjtu.edu.cn, zhaohai@cs.sjtu.edu.cn, liulongxiang21s@ict.ac.cn. Z. Zhang and L. Liu contribute equally to this work. (Corresponding author: Hai Zhao)}
\thanks{This work was partially supported by Key Projects of National Natural Science Foundation of China (U1836222 and 61733011).}
\thanks{Part of this study has been accepted as "Filling the Gap of Utterance-aware and Speaker-aware Representation for Multi-turn Dialogue" \cite{liumdfn} in the Thirty-Fifth AAAI Conference on Artificial Intelligence (AAAI 2021), {with partially material overlapped. This article extends the conference version by studying channel-aware decoupling in a broader view of multi-turn dialogue modeling. Towards this goal, we extend the descriptions in Introduction, Related Work, Model, Experiments, and Analysis correspondingly. For the techniques, this work extends the proposed model with domain-adaptive strategies, more baselines, and comprehensive analyses with new conclusions.}}}

\maketitle

\begin{abstract}
Training machines to understand natural language and interact with humans is one of the major goals of artificial intelligence. Recent years have witnessed an evolution from matching networks to pre-trained language models (PrLMs). In contrast to the plain-text modeling as the focus of the PrLMs, dialogue texts involve multiple speakers and reflect special characteristics such as topic transitions and structure dependencies between distant utterances. However, the related PrLM models commonly represent dialogues sequentially by processing the pairwise dialogue history as a whole. Thus the hierarchical information on either utterance interrelation or speaker roles coupled in such representations is not well addressed. In this work, we propose compositional learning for holistic interaction across the utterances beyond the sequential contextualization from PrLMs, in order to capture the utterance-aware and speaker-aware representations entailed in a dialogue history. We decouple the contextualized word representations by masking mechanisms in Transformer-based PrLM, making each word only focus on the words in current utterance, other utterances, and two speaker roles (i.e., utterances of sender and utterances of the receiver), respectively. In addition, we employ domain-adaptive training strategies to help the model adapt to the dialogue domains. Experimental results show that our method substantially boosts the strong PrLM baselines in four public benchmark datasets, achieving new state-of-the-art performance over previous methods. 
\end{abstract}

\begin{IEEEkeywords}
Dialogue Modeling, Open Domain Conversation System, Natural Language Generation, Deep Neural Networks.
\end{IEEEkeywords}

%
\IEEEpeerreviewmaketitle

\section{Introduction}\label{sec:introduction}
%
%
%
%
\IEEEPARstart{L}anguage is not only an effective medium for people to communicate with each other but also a natural interface between humans and machines. However, building an intelligent dialogue system that can understand human conversations and give logically correct, fluent responses is one of the eternal goals of artificial intelligence. It has been drawing increasing interest from both academia and industry areas. The methods of building a chatbot that is capable of performing multi-turn dialogue can be categorized into two lines: generation-based methods and retrieval-based methods. Generation-based methods \cite{shang2015neural,serban2015building,su2019improving,zhang2020neural,li2019data,zhao-etal-2020-knowledge-grounded,shen-feng-2020-cdl,9552464,9525043,9723498} directly generate a response using an encoder-decoder framework, which tends to be short and lacks diversity. Retrieval-based methods \cite{wu2016sequential,tao2019multi,tao2019one,zhang2018modeling,zhou2018multi,yuan2019multi,li2021deep,hua2020learning} retrieve a list of response candidates, then use a model to rank the candidates and select the best one as a reply. Since the responses of retrieval-based are generally more natural, fluent, and syntactically correct, retrieval-based methods are more mature for producing multi-turn dialogue systems both in academia and industry \cite{shum2018eliza,zhu-etal-2018-lingke,li2017alime}, which is our major focus in this work. Table \ref{tab:mutual} shows an example from Multi-Turn Dialogue Reasoning dataset (MuTual) \cite{cui2020mutual}. In order to choose the right answer, the machine is required to understand and infer from the meaning of \textit{"table"} and its coreference, indicating the requirement of reasoning ability instead of simple matching.

\begin{table}
	\centering
	\caption{\label{tab:mutual} An example of response-selection for multi-turn dialogue in MuTual dataset. \textit{F} and \textit{M} denote different speakers.
    }
	{
		\begin{tabular}{l}
		    \toprule
		     \textbf{Utterance (Context)} \\
			 \textit{F:}  \textit{Excuse me, sir. This is a non smoking area.} \\
			 \textit{M:}   \textit{Oh, sorry. \underline{I} will move to the smoking area.} \\
			 \textit{F:}  \textit{I’m afraid no \underline{table} in the smoking area is} \textit{available now.}\\
			\midrule 
			 \textbf{Response Candidates} \\
			 \textit{A.}  \textit{Sorry. I won’t smoke in the \underline{hospital} again. \xmark} \\
			\textit{B.}    \textit{OK. I won't smoke. Could you please give me} \textit{a \underline{menu}? \cmark} \\
			\textit{C.}    \textit{Could you please tell the \underline{customer over there} not to \underline{smoke}?} \\
			\quad \textit{We can't stand the smell \xmark} \\
			\textit{D.}   \textit{Sorry. I will smoke when I get off the \underline{bus}. \xmark}\\
			\bottomrule
		\end{tabular}
	}
	
\end{table}

Early studies concerning dialogue comprehension mainly focus on the matching networks {{that calculate}} similarity scores between the pairwise sequence of dialogue context and candidate response  at different granularities. The matching matrices will be fused to get a feature vector, then the sequence of feature vectors will be further integrated by RNNs to get the final representation for scoring. However, these methods have two sides of disadvantages. First, interactions mainly involve each utterance and response, ignoring the global interactions between utterances. Second, the relative positions between the response and different utterances are not taken into consideration, lacking the sequential information of context-response pairs.

Recently, pre-trained language models (PrLMs), such as BERT \cite{devlin2018bert}, RoBERTa \cite{liu2019roberta}, and ELECTRA \cite{clark2020electra}, have achieved impressive performance in a wide range of NLP tasks \cite{zhang2020mrc,zhang2019explicit,zhang2020SemBERT,zhang2019sgnet,li2020explicit,li2020global,li2019dependency}. The word embeddings derived by these language models are pre-trained on large corpora and are then fine-tuned in task-specific datasets. In the multi-turn dialogue scope, there are also some efforts using PrLMs to yield promising performances \cite{whang2019domain,henderson2019training}. The mainstream is employing PrLMs as encoders by concatenating the context and candidate responses directly, as a linear sequence of successive tokens and implicitly capturing the contextualized representations of those tokens through self-attention \cite{qu2019bert,liu2020hisbert,gu2020speaker,xu2021learning}. However, simply embedding the token to high-dimensional space cannot faithfully model the dialogue-related information, such as positional or turn order information \cite{wang2019encoding}; In addition, the mechanism of self-attention runs through the whole dialogue, resulting in entangled information that originally belongs to different parts. 


Multi-turn dialogue modeling has critical challenges of transition and inherency, which the existing multi-turn dialogue methods rarely consider to our best knowledge.

1) There exists speaker role switching in multi-turn dialogue. Each speaker has different thinking habits and speaking purposes, which lead to the unique speaking style of each speaker's role and fuzziness of utterance coreference. Therefore, it is the essential difference between dialogue and passage. Although Speaker-Aware BERT \cite{gu2020speaker} has considered role transitions, adding the embedding of speaker-id into the inputs of PrLMs is not effective enough. 

2) Utterances have their own inherent meaning and contextual meaning. A clear understanding of local and global meaning can reduce the negative influence of irrelevant content in a long-distance context. Some studies on neural machine translation have taken this into account and achieved gratifying results\cite{zheng2020toward}, {{which inspire this work.}}

{3) The widely adopted PrLMs are pre-trained on general corpora, which would be less effective
to directly fine-tune these models on downstream tasks if there is a domain shift. A recent trend is to post-train PrLMs on dialogue corpus and achieve state-of-the-art results \cite{whang2020domain,gu2020speaker,xu2021learning}. We also employ this
post-training method in this work and investigate whether it can be enhanced by modeling the dialogue properties discussed above.}

In this work, beyond the sequential contextualization from PrLMs, we proposed a novel end-to-end {{Channel-aware Decoupling Network (CDN)}} to enhance dialogue comprehension with compositional utterance interaction and thus fill the obvious gap of utterance-aware and speaker-aware representations. In detail, the contextualized word representation of dialogue text is decomposed into four parts that contain different information and then fused after sufficient interactions. More specifically, the PrLM receives concatenated context and response and outputs the contextualized representation of each word. For each word, we use a masking mechanism inside a self-attention network to limit the focus of each word only to the words in current utterance, other utterances, utterances of the sender, and utterances of the receiver respectively. To avoid ambiguity, we call the first two complementary parts the utterance-aware channel and the last two the speaker-aware channel. We then fuse the information inside each channel via a gating mechanism to control the information reservation. The word-level information will be further aggregated to the utterance level. For utterance-level representations, BiGRU is adopted to get dialogue-level representations. Inspired by \citet{tao2019multi}, we put the information fusion of two channels at the end to get the final dialogue representation for classification. In addition, we employ domain-adaptive training strategies to help the model adapt to the dialogue domains. Experimental results on four public benchmark datasets show that the proposed model outperforms the baseline models substantially on all the evaluation metrics and achieves new state-of-the-art results.

{In summary, our contributions are mainly in three folds:}

\begin{enumerate}
\item {A novel end-to-end Mask-based Decoupling-Fusion Network is proposed to fill the gap between utterance-aware and speaker-aware representations for multi-turn dialogue. }

\item {An effective domain-adaptive post-training method with  intra-utterance and inter-utterance objectives to better adapt PrLMs to our dialogue
comprehension tasks.}

\item {Experimental results on public datasets show the superiority of our method and verify the effectiveness of our model beyond strong PrLMs. We have also achieved various state-of-the-art performances compared with existing methods.} 

\end{enumerate}

\begin{figure*}[!htb]
\centering
\includegraphics[width=1.0\linewidth]{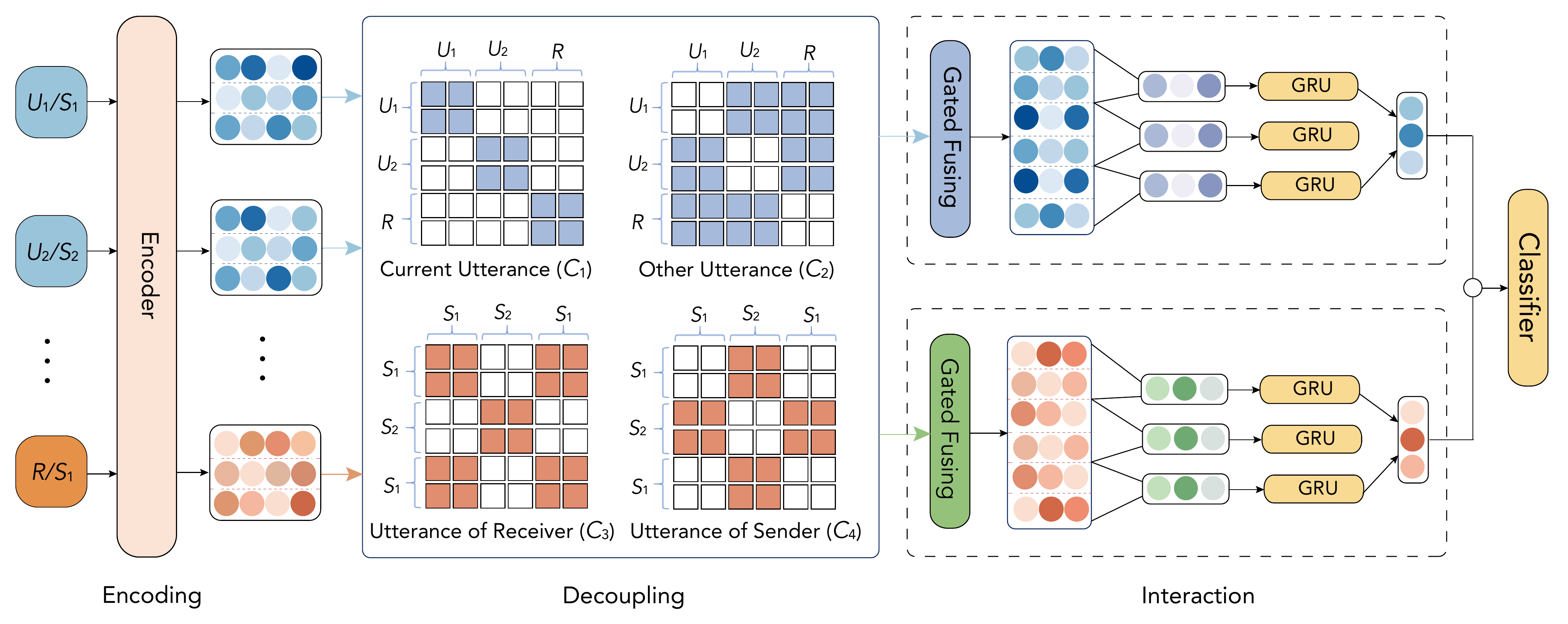} 
\caption{{Overall framework of CDN. Here for simplicity, inputs contain two utterances and one response spoken by two speakers with total six words. A more detailed figure and description on the decoupling block will be shown below. The decoupling block is composed of four independent self-attention blocks with the same inputs and different masks. The same word in different blocks attends to different scope of information.}}
\label{fig:overall}
\end{figure*}
\section{Background and Related Work}
\subsection{Multi-turn Dialogue Comprehension}
As an active research topic in the NLP field, multi-turn dialogue comprehension has attracted great attention from both academia and industry whose aim is to teach machines to read dialogue contexts and solve typical tasks such as response selection \cite{lowe2015ubuntu,wu2016sequential,zhang2018modeling,cui2020mutual,9025776}. However, selecting a coherent and informative response for a given dialogue context remains a challenge. The multi-turn dialogue typically involves two or more speakers that engage in various conversation topics and intentions. Thus the utterances are rich in interactions, e.g., with criss-cross discourse structures \cite{li2020molweni,liu2021coreference,liu2019reading}. A critical challenge is learning rich and robust context representations and interactive relationships of dialogue utterances so that the resulting model can adequately capture the semantics of each utterance and the relationships among all the utterances inside the dialogue. Previous studies can be classified into two lines: matching models and pre-trained language models.

\subsection{Matching Models}
\label{sec:relatedwork}
Matching models aim to match the response with contexts and calculate the matching score. They can be divided into two categories: single-turn and multi-turn matching models. Earlier research mainly considered the context utterances as one single utterance, using the dual encoder framework to encode the whole context and the response, respectively. The encoder varies in LSTM \cite{hochreiter1997long}, CNNs \cite{lecun1998gradient}, Attentive-LSTM \cite{tan2015lstm}.


As for multi-turn matching, Sequential Matching Network (SMN) \cite{wu2016sequential} is the representative work, where a word-word and a sequence-sequence similarity matrix will be calculated, forming a matching image for each utterance-response pair which will be further integrated by CNN and RNN. Several related works like Deep Utterance Aggregation (DUA) \cite{zhang2018modeling}, Deep Attention Matching Network (DAM) \cite{zhou2018multi}, Interaction-over-Interaction (IoI) \cite{tao2019one} are extensions of SMN from different perspectives. DUA \cite{zhang2018modeling} points out that the last utterance should be considered explicitly and use self-attention to get a better segment matching matrix. DAM \cite{zhou2018multi} uses hierarchically stacked layers of self-attention to represent matching from different levels. IoI \cite{tao2019one} mainly deepens the layers of response-utterance interaction blocks and combines final states at different depths. 

\subsection{Pre-trained Language Models}
Recently, inspired by the impressive performance of PrLMs, the mainstream is employing PrLMs to handle the whole pairwise texts as a linear sequence of successive tokens and implicitly capture the contextualized representations of those tokens through self-attention \cite{qu2019bert,liu2020hisbert,gu2020speaker,xu2021learning}. The word embeddings derived by these language models are pre-trained on large corpora and are then utilized as either distributed word embeddings \cite{peters2018deep} or fine-tuned according to the specific task needs \cite{devlin2018bert}. 
Most of the pre-trained Language Models (PrLMs) are based on Transformer, among which Bidirectional Encoder Representations from Transformers (BERT) \cite{devlin2018bert} is one of the most representative works. BERT uses multiple layers of stacked Transformer Encoder to obtain contextualized representations of the language at different levels. BERT has achieved unprecedented performances in many downstream tasks of NLP. Several subsequent variants have been proposed to further enhance the capacity of PrLMs, such as RoBERTa \cite{liu2019roberta}, ALBERT \cite{lan2019albert}, and ELECTRA \cite{clark2020electra}. 
To evaluate our methods on both sides of the fast inference speed and strong performance, we will adopt BERT and ELECTRA as our backbone models in this work.

\section{Channel-aware Decoupling Network}
Our proposed model, named Channel-aware Decoupling Network (CDN), consists of six parts: Encoding, Decoupling, Fusing, Words Aggregating, Utterances Integrating, and Scoring. In this section, we will formulate the problem and then introduce each sub-module in detail.
\subsection{Problem Formulation}
Suppose that we have a dataset $\mathcal{D} = \{(y_i, c_i, r_i)\}^N_{i=1}$, where $c_i = \{u_{i,1},...,u_{i,n_i}\}$ represents the dialogue context with $\{u_{i,k}\}^{n_i}_{k=1}$ as utterances. $r_i$ is a response candidate, and $y_i \in \{0,1\}$ denotes a label. The goal is to learn a discriminator $g(\cdot,\cdot)$ from $\mathcal{D}$, and at the inference phase, given any new context $c$ and response $r$, we use the discriminator to calculate $g(c,r)$ as their matching score. To select several best responses as return to human inputs, we rank matching scores or set a threshold among the list of candidates.
\subsection{Model Overview}

Figure \ref{fig:overall} shows the overall framework of CDN. CDN first encodes each word in the concatenated context and response to contextualized representation using the PrLMs. In the decoupling module, a self-attention mechanism with different masks forces each word to only focus on the information in specific scopes, decoupling to utterance-aware and speaker-aware channels. Note that now the information flow inside different channels is independent. In each channel, complementary information will be fused by a gate. Then word representations belonging to the same utterance will be aggregated to represent the utterance. The sequence of utterance representations will be delivered to Bidirectional Gated Recurrent Unit (BiGRU) to produce the channel-aware dialogue-level representation. After that, the information from different channels is fused to get the final dialogue representation. The classifier will then output a score of the dialogue to reflect the matching degree between context and candidate response. 

CDN is superior to existing methods in the following ways. First, compared to SA-BERT \cite{gu2020speaker}, the speaker-aware information is built upon semantically rich contextualized representations instead of a simple additional embedding term, making its influence on dialogue scoring preserved better. Second, compared to the work of designing a complicated matching network, our structure is simpler and can represent information at different levels in a more unified manner. Third, compared to pure PrLMs, the additional layer of CDN can decouple information to different channels, making full use of the contextualized representations. Besides, since the number of utterances is far less than that of words, the top BiGRU layers can keep their memory capacity, alleviating the defects of the Transformer in reflecting relative position information.  
\subsection{Encoding Context and Response}
\label{sec:PrLMs}
We employ a pre-trained language model such as ELECTRA to obtain the initial word representations. {{The tokens in utterances and response are concatenated successively and then fed into the encoder}. Following \citet{devlin2018bert}, we add ``[CLS]" token in the front of the sequence and insert ``[SEP]" between adjacent utterances.} Let $U_i = [u_{i,0}, u_{i,1}, ..., u_{i,n_i}]$, $R = [r_0, r_1, ..., r_{n_r}]$, where $n_i$ and $n_r$ is respectively the length of $i$-th utterances and response, then the output is $ E = [e_1, e_2, ..., e_{n_0 + ... + n_k + n_r}]$ where $k$ is the maximum number of utterances, and $e_i \in \mathbf{R}^d$, $d$ is hidden size.
\subsection{Channel-aware Information Decoupling}
We first introduce the multi-head self-attention (MHSA) with mask, which can be formulated as:
\begin{equation}
\begin{split}
    & \text{Attention}(Q,K,V,M) = \text{softmax}(\dfrac{QK^T}{\sqrt{d_k}} + M)V, \\ 
    & \text{head}_i = \text{Attention}(EW_i^Q,EW_i^K,EW_i^V,M),\\
    & \text{MHSA}(E,M) = \text{Concat}(\text{head}_1,...,\text{head}_h)W^O,
\end{split}
\end{equation}
where $W_i^Q \in \mathbb{R}^{d_{model}\times d_q}$, $W_i^K \in \mathbb{R}^{d_{model} \times d_k}$, $W_i^V \in \mathbb{R}^{d_{model} \times d_v}$, $W_i^O \in \mathbb{R}^{hd_v \times d_{model}}$ are parameter matrices, $d_q$, $d_k$, $d_v$ denote the dimension of Query vectors, Key vectors and Value vectors, $h$ denotes the number of heads. $M$ denotes the mask.

We have four masks $\{M_k\}_{k=1}^4 \in \mathbb{R}^{l \times l}$ defined as:
\begin{equation}
    \begin{split}
        M_1[i, j] &=\left\{\begin{array}{cc}
0, & \text { if } \mathbb{T}_{i} = \mathbb{T}_{j} \\
-\infty, & \text { otherwise }
\end{array}\right.\\
        M_2[i, j] &=\left\{\begin{array}{cc}
0, & \text { if } \mathbb{T}_{i} \neq \mathbb{T}_{j} \\
-\infty, & \text { otherwise }
\end{array}\right.\\
        M_3[i, j] &=\left\{\begin{array}{cc}
0, & \text { if } \mathbb{S}_{i} = \mathbb{S}_{j} \\
-\infty, & \text { otherwise }
\end{array}\right.\\
        M_4[i, j] &=\left\{\begin{array}{cc}
0, & \text { if } \mathbb{S}_{i} \neq \mathbb{S}_{j} \\
-\infty, & \text { otherwise }
\end{array}\right.\\
    \end{split}
\end{equation}
where $i$, $j$ is the word position in the whole dialogue, $\mathbb{T}_i$ is the index of utterance the $i$-th word is located in, and $\mathbb{S}_i$ is the speaker that the $i$-th word is spoken by. And $M_1$, $M_2$, $M_3$, $M_4$ only attend to the word in current utterance, other utterances, utterances of sender, and utterances of receiver. We call the first two utterance-aware channels and the last two speaker-aware channels.

The decoupled channel-aware information $\{C_k\}_{k=1}^4 \in \mathbb{R}^{l \times d}$ are derived by multi-head self-attention with mask. 
\begin{equation}
\label{eq:CI}
    C_i = \text{MHSA}(E, M_i), i\in \{1,2,3,4\},
\end{equation}
where $E \in \mathbb{R}^{l \times d}$ is the output of PrLMs stated in Section \ref{sec:PrLMs}.
\subsection{Complementary Information Fusing}
To fuse the complementary information inside each channel, we use a gate to fuse them. Inspired by \cite{mou2015natural}, we use a gate to calculate the ratio of information preservation based on matching heuristics considering the ``Siamese'' architecture as information from two parts, the element-wise product and difference as “similarity” or “closeness” measures. Let $E$ denote the original representations output from PrLMs. For utterance-aware channel, $\bar{E} = C_1$ and $\hat{E} = C_2$. For speaker-aware channel, $\bar{E} = C_3$ and $\hat{E} = C_4$. $C_i$ is defined in Equation (\ref{eq:CI}). The gate is formulated as:
\begin{equation}
    \label{eq:gate}
    \begin{split}
        &\Tilde{E_1} = \text{ReLU}(\text{FC}([E,\bar{E}, E-\bar{E}, E \odot \bar{E}])), \\
        &\Tilde{E_2} = \text{ReLU}(\text{FC}([E,\hat{E}, E-\hat{E}, E \odot \hat{E}])), \\
        &P = \text{Sigmoid}(\text{FC}(([\Tilde{E_1}, \Tilde{E_2}]))), \\
        &G(E, \bar{E}, \hat{E}) = P, \\
    \end{split}
\end{equation}
where Sigmoid, ReLU \cite{agarap2018deep} are activation functions, FC is fully-connected layer, [$\cdot,\cdot$] means concatenation, {and $\odot$ is element-wise multiplication}. 

Using two parametric-independent gates, the channel-aware information can be fused, which is defined as:
\begin{equation}
    \begin{split}
        & P_1 = G_1(E,C_1,C_2), \\
        & P_2 = G_2(E,C_3,C_4), \\
        & C_u = P_1 \odot C_1 + (1-P_1)  \odot C_2, \\
        & C_s = P_2 \odot C_3 + (1-P_2)  \odot C_4,
    \end{split}
\end{equation}
where the calculation of $\{G_i\}_{i=1}^2$ is defined in Equation \ref{eq:gate}, $C_u$ and $C_s$ is the fused utterance-aware and speaker-aware word representations, respectively.
\subsection{Utterance Representations}
For each channel, word representations will be aggregated by simple max-pooling over words in the same utterance to get the utterance representations. Let $L_u$ and $L_s$ be the output in this part. Then they are defined as:
\begin{equation}
\begin{split}
    & L_u[i,:] = \mathop{\text{MaxPooling}}\limits_{\mathbb{T}_j=i}({C_u[j,:]}) \in \mathbb{R}^d, \\
    & L_s[i,:] = \mathop{\text{MaxPooling}}\limits_{\mathbb{T}_j=i}({C_s[j,:]}) \in \mathbb{R}^d.
\end{split}
\label{eq:l}
\end{equation}
\subsection{Dialogue Representation}
To get the channel-aware dialogue representation, the sequence of utterance representations will be delivered to BiGRU. Suppose that the hidden states of the BiGRU are $(\boldsymbol{h}_1, \dots, \boldsymbol{h}_k)$, then $\forall j$, $1\leq j \leq k$, $\boldsymbol{h}_j \in \mathbb{R}^{2d}$ is given by
\begin{equation}
    \begin{split}
        &\overleftarrow{\boldsymbol{h}}_j = \overleftarrow{\text{GRU}}(\overleftarrow{\boldsymbol{h}}_{j-1}, \overleftarrow{\boldsymbol{L}}[j]) ,\\
        &\overrightarrow{\boldsymbol{h}}_j = \overrightarrow{\text{GRU}}(\overrightarrow{\boldsymbol{h}}_{j-1}, \overrightarrow{\boldsymbol{L}}[j]) ,\\
        & \boldsymbol{h}_j = [\overleftarrow{\boldsymbol{h}}_j; \overrightarrow{\boldsymbol{h}}_j].
    \end{split}
\end{equation}
{where $\boldsymbol{L}[j]$ denotes the sentence representation derived from Eq. (\ref{eq:l}).} For each channel, we take the hidden state of BiGRU at the last step as channel-aware dialogue representation. Let the two vectors be $\boldsymbol{v_1}$ and $\boldsymbol{v_2}$. Then two channels can be easily fused to get the final dialogue representation.
\begin{equation}
    \boldsymbol{v} = \text{Tanh}(W[\boldsymbol{v_1};\boldsymbol{v_2}] + b),
\end{equation}
where $W \in \mathbb{R}^{d \times 4d}$, $b \in \mathbb{R}^{d}$ are trainable parameters. Tanh is the activation function.
\subsection{Scoring}
The dialogue vector will be fed into a classifier with a fully connected and softmax layer. We learn model $g(\cdot, \cdot)$ by minimizing cross-entropy loss with dataset $\mathcal{D}$. Let $\Theta$ denote the parameters of CDN. For binary classification like ECD, Douban, and Ubuntu, the objective function $\mathcal{L(D}, \Theta)$ can be formulated as:
\begin{equation}
    -\sum_{i=1}^N [y_ilog(g(c_i,r_i)) + (1-y_i)log(1-g(c_i,r_i))].
\end{equation}
{where $N$ is the number of training data}. For multiple-choice tasks like MuTual, the loss function is:
\begin{equation}
    -\sum_{i=1}^N\sum_{k=1}^C y_{i,k}log(g(c_i,r_{i,k})).
\end{equation}
{where $C$ is the number of candidate response options for each input context.}

\section{Domain-Adaptive Post-Training}
Although the PrLMs demonstrate superior performance due to their strong representation ability from self-supervised pre-training, it is still challenging to effectively adapt task-related knowledge during the detailed task-specific training, which is usually in the way of fine-tuning \cite{gururangan-etal-2020-dont}. Generally, those PrLMs handle the whole input text as a linear sequence of successive tokens and implicitly capture the contextualized representations of those tokens through self-attention. The such fine-tuning paradigm of exploiting PrLMs would be suboptimal to model dialogue tasks that hold exclusive text features that plain text for PrLM training may hardly embody. In this following section, we first describe the standard general-purpose training in BERT, and then present our extensions to the concerned dialogue domains.

\subsection{General-Purpose Training} 
As the standard pre-training procedure, PrLMs are pre-trained on large-scale domain-free texts and then used for fine-tuning according to the specific task needs. There are token-level and sentence-level objectives used in the general-purpose pre-training. The most widely-used PrLM for domain-adaption in the dialogue field is BERT \cite{devlin2018bert}, whose pre-training is based on two loss functions: (1) a masked language model (MLM) loss, and (2) a next sentence prediction (NSP) loss. MLM first masks out some tokens from the input sentences and then trains the model to predict them by the rest of the tokens. NSP is another widely used pre-training objective, which trains the model to distinguish whether two input sentences are continuous segments from the training corpus. 
\subsection{Domain-Adaptive Training}
The original PrLMs are trained on a large text corpus to learn general language representations. To incorporate specific in-domain knowledge, adaptation on in-domain corpora, also known as domain-aware pre-training, is designed, which directly employs the original PrLMs as mentioned in the general-purpose paragraph above, using the dialogue-domain corpus. In this work, we extend the MLM and NSP as intra-utterance and inter-utterance objectives to coordinate with our dialogue comprehension tasks.

$\bullet$ \textbf{Intra-utterance Objective} To better adapt PrLMs to dialogue scenarios, we conduct the post-training with three levels of masks for MLM, including subword, word, and span-level masks. \textit{Subword} is the original method used in BERT \cite{devlin2018bert}, and \textit{word} means the {whole-word-masking (WWM)} where the whole word will be masked during the data preprocessing. {{The \textit{Span} masking mechanism follows \cite{joshi2020spanbert}, which masks contiguous random spans from a geometric distribution rather than random tokens.}} Compared with subword masks, both span and {WWM} mechanisms would potentially improve the model's ability to capture high-level topic information in dialogue texts. The training objective is then formed as follows:
\begin{equation}
\begin{split}
\mathbb{L}_{intra} &= -\sum_{k=1}^{N} \left [ {m}_{i}\log\hat{m}_{i} \right ],
\end{split}
\end{equation}
where $\hat{m}_{i}$ denotes the predicted token id and ${m}_{i}$ is the ground-truth. $N$ is the number of examples.

$\bullet$ \textbf{Inter-utterance Objective}
For a dialogue context with multiple utterances $c= \{u_{1}, ... , u_{n}\}$, we form a context text $c'=\{u_{1}, ..., u_{k-1}\}, k\in[1,n]$ for each utterance $u^{+}_k$. We then randomly sample an utterance $u^{-}_k$ from the corpus as the negative example. The goal is to identify utterance $u_k$ as the true utterance against the negative one by feeding sequence ``[CLS] $c'$ [SEP] $u_k$ [SEP]'' into BERT. Following the setting proposed in BERT, we take the hidden representation of [CLS] through a full-connected layer followed by a sigmoid function for classification. Here, we minimize the cross entropy loss in {post-training}.
\begin{equation}
\begin{split}
\mathbb{L}_{inter} &= -\sum_{k=1}^{N} \left [ {s}_{i}\log\hat{s}_{i} \right ],
\end{split}
\end{equation}
where $\hat{s}_{i}$ denotes the predicted label and ${s}_{i}$ is the ground-truth. $N$ is the number of examples.

During post-training, we combine both of the intra- and inter- utterance objectives:
\begin{equation}
    \mathbb{L} = \mathbb{L}_{intra} + \mathbb{L}_{inter}.
\end{equation}

\begin{table*}[t]
\centering
 \caption{Test results on Ubuntu, Douban, and E-commerce datasets. $\dagger$ denotes the methods with domain-adaptive post-training.
   The evaluation results are collected from published literature \cite{zhang2021kkt,whang2019domain,xu2021learning,lin2020world}. 
  } 
  {\begin{tabular}{lc c cc c c c c cc c c}
    \toprule
    \multirow{2}{*}{\textbf{Models}} & \multicolumn{3}{c}{\textbf{Ubuntu}} & \multicolumn{6}{c}{\textbf{Douban}} & \multicolumn{3}{c}{\textbf{E-commerce}}\\
      & $\textbf{R}_{10}$@1 & $\textbf{R}_{10}$@2 & $\textbf{R}_{10}$@5 & \textbf{MAP} & \textbf{MRR} & \textbf{P}@1 & $\textbf{R}_{10}$@1 & $\textbf{R}_{10}$@2 & $\textbf{R}_{10}$@5 & $\textbf{R}_{10}$@1 & $\textbf{R}_{10}$@2 & $\textbf{R}_{10}$@5 \\
    \midrule
    \multicolumn{13}{l}{\textit{Single-turn models with concatenated matching}}\\
    CNN \cite{kadlec2015improved} & 0.549 & 0.684 & 0.896 & 0.417 & 0.440 & 0.226 & 0.121 & 0.252 & 0.647 & 0.328 & 0.515 & 0.792\\
    LSTM \cite{kadlec2015improved} & 0.638 & 0.784 & 0.949 & 0.485 & 0.537 & 0.320 & 0.187 & 0.343 & 0.720 & 0.365 & 0.536 & 0.828\\
    BiLSTM \cite{kadlec2015improved} & 0.630 & 0.780 & 0.944 & 0.479 & 0.514 & 0.313 & 0.184 & 0.330 & 0.716 & 0.365 & 0.536 & 0.825\\
    MV-LSTM \cite{wan2016match} & 0.653 & 0.804 & 0.946 & 0.498 & 0.538 & 0.348 & 0.202 & 0.351 & 0.710 & 0.412 & 0.591 & 0.857\\ 
    Match-LSTM \cite{wang2016learning} & 0.653 & 0.799 & 0.944 & 0.500 & 0.537 & 0.345 & 0.202 & 0.348 & 0.720 & 0.410 & 0.590 & 0.858\\ 
    \midrule
    \multicolumn{13}{l}{\textit{Multi-turn matching network with separate interaction}}\\
    Multi-View \cite{zhou2016multi} & 0.662 & 0.801 & 0.951 & 0.505 & 0.543 & 0.342 & 0.202 & 0.350 & 0.729 & 0.421 & 0.601 & 0.861\\
    DL2R \cite{yan2016dl2r} & 0.626 & 0.783 & 0.944 & 0.488 & 0.527 & 0.330 & 0.193 & 0.342 & 0.705 & 0.399 & 0.571 & 0.842\\
    SMN \cite{wu2016sequential} & 0.726 & 0.847 & 0.961 & 0.529 & 0.569 & 0.397 & 0.233 & 0.396 & 0.724 & 0.453 & 0.654 & 0.886\\
    DUA \cite{zhang2018modeling} & 0.752 & 0.868 & 0.962 & 0.551 & 0.599 & 0.421 & 0.243 & 0.421 & 0.780 & 0.501 & 0.700 & 0.921\\
    DAM \cite{zhou2018multi} & 0.767 & 0.874 & 0.969 & 0.550 & 0.601 & 0.427 & 0.254 & 0.410 & 0.757 & 0.526 & 0.727 & 0.933\\
    MRFN \cite{tao2019multi} & 0.786 & 0.886 & 0.976 & 0.571 & 0.617 & 0.448 & 0.276 & 0.435 & 0.783 & - & - & - \\
    IMN \cite{gu2019interactive} & 0.794 & 0.889 & 0.974 & 0.570 & 0.615 & 0.433 & 0.262 & 0.452 & 0.789 & 0.621 & 0.797 & 0.964 \\
    IoI \cite{tao2019one} & 0.796 & 0.894 & 0.974 & 0.573 & 0.621 & 0.444 & 0.269 & 0.451 & 0.786 & 0.563 & 0.768 & 0.950\\
    MSN \cite{yuan2019multi} & 0.800 & 0.899 & 0.978 & 0.587 & 0.632 & 0.470 & 0.295 & 0.452 & 0.788 & 0.606 & 0.770 & 0.937\\
    G-MSN \cite{lin2020world} & 0.812 & 0.911 & 0.987 & 0.599 & 0.645 & 0.476 & 0.308 & 0.468 & 0.826 & 0.613 & 0.786 & 0.964 \\
    \midrule
    \multicolumn{13}{l}{\textit{PrLM-based methods for fine-tuning}}\\
    
    BERT-SS-DA \cite{lu2020improving} & 0.813 & 0.901 & 0.977 & 0.602 & 0.643 & 0.458 & 0.280 & 0.491 & 0.843 & 0.648 & 0.843 & 0.980\\ 
    TADAM \cite{xu2021topic}  & 0.821 & 0.906 & 0.978 & 0.594 & 0.633 & 0.453 & 0.282 & 0.472 & 0.828 & 0.660 & 0.834 & 0.975 \\
    PoDS \cite{zhang2021kkt}  & 0.828  &  0.912 & 0.981  & 0.598  & 0.636  & 0.460  &  0.287 & 0.468  &  0.845 &  0.633 & 0.810  &   0.967  \\
    ELECTRA \cite{liumdfn} & 0.845 & 0.919 & 0.979 & 0.599 & 0.643 & 0.471 & 0.287 & 0.474 & 0.831 & 0.607 & 0.813 & 0.960 \\
    SA-BERT$\dagger$ \cite{gu2020speaker} & 0.855 & 0.928 & 0.983 & 0.619 & 0.659 & 0.496 & 0.313 & 0.481 & 0.847 & 0.704 & 0.879 & 0.985\\
    PoDS$\dagger$ \cite{zhang2021kkt}  & {0.856} & {0.929} & {0.985} &  {0.599}  & {0.637}  & 0.460  &  0.287 &  {0.469} & 0.839 &  {0.671} & {0.842}  &   {0.973}  \\
    BERT-VFT$\dagger$ \cite{whang2020domain} & 0.858 & 0.931 & 0.985 & - & - & - & - & - & -  & - & - & -  \\
    DCM$\dagger$ \cite{li2021deep} & 0.868 & 0.936 & 0.987 & 0.611 & 0.649 & - & 0.294 & 0.498 & 0.842 & 0.685 & 0.864 & 0.982 \\
    {{{UMS$_{\text{ELECTRA}}$\cite{whang2021response}}}} & {0.854} & 0.929 & {0.984} & 0.608 & 0.650 & 0.472 & 0.291 & 0.488 & 0.845 & 0.648 & 0.831 & 0.974 \\
    {{{UMS$_{\text{ELECTRA}}\dagger$ \cite{whang2021response}}}} & {0.875} & 0.941 & {0.988} & 0.623 & 0.663 & 0.492 & 0.307 & 0.501 & 0.851 & 0.707 & 0.853 & 0.974\\
    \midrule
    \multicolumn{13}{l}{\textit{Our Implementation}}\\
    ELECTRA & 0.845 & 0.919 & 0.979 & 0.599 & 0.643 & 0.471 & 0.287 & 0.474 & 0.831 & 0.607 & 0.813 & 0.960 \\
    {\quad + CDN} & 0.866 & {0.932} & {0.984} &  {0.624} & {0.663} & {0.498} & {0.325} & {0.511} & {0.855} & 0.639 &  0.829 & 0.971 \\
    \hdashline
    ELECTRA$\dagger$ & 0.890  & 0.947  & 0.989 &  0.625 &	0.663 &	0.483 &	0.301 &	0.513 &	0.865  & 0.657 & 0.834 & 0.977 \\
    {\quad + CDN}$\dagger$ &  \textbf{0.914} & \textbf{0.961}  & \textbf{0.993} & \textbf{0.634} & \textbf{0.674}  & \textbf{0.496} & \textbf{0.312} & 	\textbf{0.540} &	\textbf{0.868}   & \textbf{0.673} & \textbf{0.857} & \textbf{0.979}  \\
  \bottomrule
  \end{tabular}}
  \label{table:rs_results}
\end{table*}

\section{Experiments}
\subsection{Datasets}
We tested our model on two English datasets: Ubuntu Dialogue Corpus (Ubuntu) and Multi-Turn Dialogue Reasoning (MuTual), and two Chinese datasets: Douban Conversation Corpus (Douban) and E-commerce Dialogue Corpus (ECD).
\subsubsection{Ubuntu Dialogue Corpus} Ubuntu \cite{lowe2015ubuntu} consists of English multi-turn conversations about technical support collected from chat logs of the Ubuntu forum. The dataset contains 1 million context-response pairs, 0.5 million for validation, and 0.5 million for testing. In the training set, each context has one positive response generated by humans and one negative response sampled randomly. In the validation and test sets, for each context, there are 9 negative responses and 1 positive response. 
\subsubsection{Douban Conversation Corpus} Douban \cite{wu2016sequential} is different from Ubuntu in the following ways. First, it is an open domain task where dialogues are extracted from the Douban Group. Second, Response candidates on the test set are collected by using the last turn as the query to retrieve 10 response candidates and labeled by humans. Third, there could be more than one correct response for a context.
{{Response candidates on the test set are collected by a standard search engine \emph{Apache Lucene}\footnote{\url{http://lucene.apache.org/}}, other than negative sampling without human judgment on Ubuntu Dialogue Corpus. }}

\subsubsection{E-commerce Dialogue Corpus} ECD \cite{zhang2018modeling} dataset is extracted from conversations between customers and service staff on Taobao. It contains over 5 types of conversations based on over 20 commodities. There are also 1 million context-response pairs in the training set, 0.5 million in the validation set, and 0.5 million in the test set.
\subsubsection{Multi-Turn Dialogue Reasoning} MuTual \cite{cui2020mutual} consists of 8,860 manually annotated dialogues based on Chinese student English listening comprehension exams. For each context, there is one positive response and three negative responses. The difference compared to the above three datasets is that only MuTual is reasoning-based. There are more than 6 types of reasoning abilities reflected in MuTual.
\subsection{Setup}
For the sake of computational efficiency, the maximum number of utterances is specialized as 20. The concatenated context, response, "[CLS]" and "[SEP]" in one sample are truncated according to the "longest first" rule or padded to a certain length, which is 256 for MuTual and 384 for the other three datasets. Our model is implemented using Pytorch and based on the Transformer Library. We use ELECTRA \cite{clark2020electra} as our underlying model. AdamW \cite{loshchilov2017decoupled} is used as our optimizer. The batch size is 24 for MuTual, and 64 for others. The initial learning rate is $4\times 10^{-6}$ for MuTual and $3\times 10^{-6}$ for others. We run up to 3 epochs for MuTual and 2 epochs for others and select the model that achieves the best result in the validation process. 

Our domain adaptive post-training for the corresponding response selection tasks is based on the three large-scale dialogue corpus, including Ubuntu, Douban, and ECD, respectively. {{The data is the same as the training data in fine-tuning but only used in {unsupervised pre-training}.}} Because there is no appropriate domain data for the small-scale Mutual dataset, we only report the fine-tuning results without post-training.

For the English tasks, we use the pre-trained weights \textit{
bert-base-uncased} and \textit{electra-large-discriminator} for fine-tuning; for the Chinese tasks, the weights are from \textit{
bert-base-chinese} and \textit{hfl/chinese-electra-large-discriminator}.\footnote{Those weights are available in the Transformers repo: \url{https://github.com/huggingface/transformers/}.}

\subsection{Baseline Models}
The following models are our baselines for comparison:

$\bullet$ \textbf{Single-turn matching models}:
{{Single-turn Matching Models consider the context utterances as one single utterance, using the dual encoder framework to encode the whole context and the response, respectively. The variants include}}
CNN, LSTM, BiLSTM \cite{kadlec2015improved}, MV-LSTM \cite{wan2016match}, and Match-LSTM
\cite{zhou2016multi} construct the dialog context by concatenating utterances as a long document, and match the dialog context with {a candidate response}.

$\bullet$ \textbf{Multi-turn matching models}: {{Multi-turn matching models focus on the utterance-level interaction with responses. The major difference lies in the different types of matching, including cross-attention, self-attention, etc. The typical studies include}} Sequential Matching Network (SMN) \cite{wu2016sequential}, Deep Attention Matching Network (DAM) \cite{zhou2018multi}, Deep Utterance Aggregation (DUA) \cite{zhang2018modeling}, Interaction-over-Interaction (IoI) \cite{tao2019one} which have been stated in Section \ref{sec:relatedwork}. Besides, Multi-Representation Fusion Network (MRFN) \cite{tao2019multi} matches context and response with multiple types of representations. Multi-hop Selector Network (MSN) \cite{yuan2019multi} utilizes a multi-hop selector to filter necessary utterances and match among them.

$\bullet$ \textbf{PrLMs-based models}: {{PrLMs-based models apply PrLMs to encode the dialogue history and candidate response as a whole.}} BERT \cite{devlin2018bert}, RoBERTa \cite{liu2019roberta}, ALBERT \cite{lan2019albert}, ELECTRA \cite{clark2020electra} have been stated in Section \ref{sec:relatedwork}. Besides, Option Comparison Network (OCN) \cite{ran2019option} is involved, which compares the options before matching response and contexts.\footnote{ On the leaderboard of MuTual, since some work is not publicly available, we will not introduce here.}

\subsection{Evaluation Metrics}
Following the previous work \cite{lowe2015ubuntu, wu2016sequential}, we calculate the proportion of true positive responses among the top-$k$ selected responses from the list of $n$ available candidates for one context, denoted as $\textbf{R}_n$@$k$. Besides, additional conventional metrics of information retrieval are employed on Douban: Mean Average Precision (MAP) \cite{baeza1999modern}, Mean Reciprocal Rank (MRR) \cite{voorhees1999trec}, and precision at position 1 (P@1). 

\subsection{Results}
Tables \ref{table:rs_results}-\ref{tab:mutual_result} show the {{test}} results on four datasets. 

1) Generally, the previous models based on multi-turn matching networks perform worse than simple PrLMs-based ones, illustrating the power of contextualized representations in context-sensitive dialogue modeling.

2) PrLM can perform even better when equipped with CDN, verifying the effectiveness of our model, where utterance-aware and speaker-aware information can be better exploited. 

3) {It is observed that post-training on dialogue data can substantially boost the baseline, i.e., by 4.5\% $\textbf{R}_{10}$@1 on Ubuntu. Even on the backbone baseline with domain-adaptive post-training, CDN can still yield consistent performance gains, which indicates that {the CDN architecture is generally effective even working with stronger backbone PrLM encoders after post-training. A similar phenomenon is observed in existing studies \cite{zhou2018multi,tao2019multi,gu2019interactive,tao2019one} that post-training on dialogue data brings more improvements than modifying attention architectures in Transformers. However, post-training is less efficient, which requires costly training on an additional domain-specific corpus.}}




4) Our model outperforms other models in most metrics. CDN surpasses the previous SOTA SA-BERT model on most datasets, which is augmented by extra domain adaptation strategies to conduct language model pre-training on in-domain dialogue corpus before fine-tuning on tasks. In addition, CDN also ranks the best on the MuTual leaderboard.\footnote{https://nealcly.github.io/MuTual-leaderboard/}

\begin{table}[htb]
{
    \centering
     \caption{\label{tab:mutual_result} {Results on {the MuTual test set}. {{The upper and middle blocks present the public models w/o and w/ PrLMs, respectively}}. 
    The results are collected from published literature \cite{lowe2015ubuntu,liumdfn,Lidapo}. The lower block shows our implementations on the reproduced BERT and ELECTRA baselines following the official implementations in the MuTual dataset paper \cite{cui2020mutual}.}
	}
    {
        \begin{tabular}{lccc}
            \toprule \textbf{Model} & $\textbf{R}_{4}$@1 & $\textbf{R}_{4}$@2 & \textbf{MRR} \\
            \midrule
            TF-IDF \cite{lowe2015ubuntu} & 0.279 & 0.536 & 0.542 \\
            Dual LSTM \cite{lowe2015ubuntu} & 0.260 & 0.491 & 0.743 \\
            SMN \cite{wu2016sequential} & 0.299 & 0.585 & 0.595 \\
            DAM \cite{zhou2018multi} & 0.241 & 0.465 & 0.518 \\
            \midrule
            GPT-2 \cite{radford2019language} & 0.332 & 0.602 & 0.584 \\
            \quad GPT-2-FT \cite{radford2019language} & 0.392 & 0.670 & 0.629 \\
            BERT \cite{devlin2018bert} & 0.648 & 0.847 & 0.795 \\
            RoBERTa \cite{liu2019roberta} & 0.713 & 0.892 & 0.836 \\
            \quad RoBERTa + OCN \cite{ran2019option} & 0.867 & 0.958 & 0.926 \\
            ALBERT \cite{liu2020graph} & 0.847 & 0.962 & 0.916 \\
            \quad GRN-v2 \cite{liu2020graph} & 0.915 & 0.983 & 0.954 \\
            \midrule
            {BERT} & 0.656 & 0.866 & 0.803\\
            \quad {+ CDN} & 0.692 & 0.881 & 0.823\\
            ELECTRA & 0.900 & 0.979 & 0.946 \\
            \quad + CDN & \textbf{0.916} & \textbf{0.984} & \textbf{0.956} \\
             \bottomrule
        \end{tabular}
    }
   
}
\end{table}

\section{Analysis}

		

	

\begin{figure*}[!htb]
\centering
\includegraphics[width=1.0\linewidth]{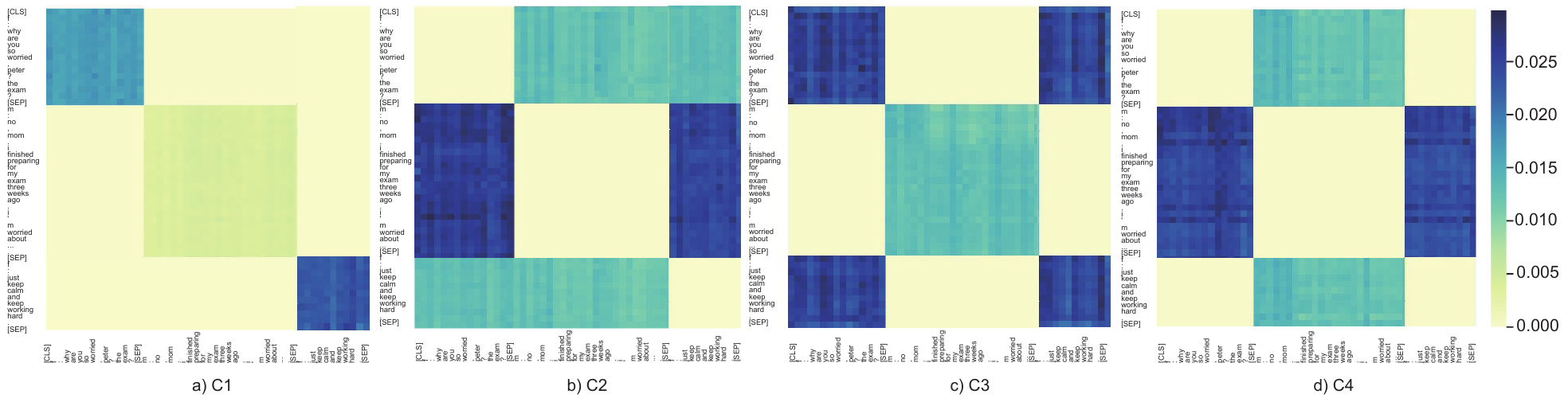} 
\caption{{Visualization of attention distribution in $C_1$, $C_2$, $C_3$ and $C_4$, which indicates that channel-aware decoupling operations distinguish the four parts of utterance-aware and speaker-aware representations. The example context is from the MuTual validation set. The text is ``\textit{[CLS] f : why are you so worried , peter ? the exam ? [SEP] m : no , mom . i finished preparing for my exam three weeks ago . i ' m worried about my paper . you know , the deadline is coming up . but i have n ' t collected enough information . if i ca n ' t finish it on time , my teacher will not be happy with me . [SEP] f : just keep calm and keep working hard . [SEP]}". The special tokens, i.e., [CLS] and [SEP] are bold in blue, which indicate the utterance boundary.}}
\label{fig:vis}
\end{figure*}

\subsection{Ablation Study}
{Since our model first decouples information to utterance-aware and speaker-aware channels and then uses BiGRU to learn a channel-aware dialogue representation from the sequence of utterance representations, we wonder about the effect of two channels and whether BiGRU can be replaced by simple pooling.
First, we visualize the attention distribution in $C_1$, $C_2$, $C_3$ and $C_4$ as shown in Figure \ref{fig:vis}. The illustration shows that channel-aware decoupling operations distinguish the four parts of utterance-aware and speaker-aware representations. }
Then, we perform an ablation study on MuTual dev sets as shown in Table \ref{tab:ablation}.\footnote{We conduct the ablation studies on MuTual because MuTual is of the highest quality, which is a manually annotated dialogues based on English listening comprehension exams.} Results show that each part is {{necessary}}. The most important part is speaker-aware information since speaker role transition is an essential feature in multi-party dialogues. Then it comes to BiGRU, revealing the strength of BiGRU {in modeling the sentence-level transition between the turns of utterances}.

\subsection{Number of Decoupling Layers}
\begin{figure}
	\centering
			\setlength{\abovecaptionskip}{0pt}
			\begin{center}
			\pgfplotsset{height=5.6cm,width=6cm,compat=1.14,every axis/.append style={thick},every axis legend/.append style={ at={(0.95,0.95)}},legend columns=3 row=1} \begin{tikzpicture} \tikzset{every node}=[font=\small] \begin{axis} [width=8cm,enlargelimits=0.13, xticklabels={1,2,3,4,5}, axis y line*=left, axis x line*=left, xtick={1,2,3,4,5}, x tick label style={rotate=0},
			ylabel={Accuracy(\%)},
			ymin=0.905,ymax=1.0,
			ylabel style={align=left},xlabel={Decoupling Block},font=\small]
			\addplot+ [smooth, mark=*,mark size=1.2pt,mark options={mark color=cyan}, color=red] coordinates
			{ (1,0.923) (2,0.909) (3,0.910) (4,0.913) (5,0.911)};
			\addlegendentry{\small R@1}
			\addplot+[smooth, mark=diamond*, mark size=1.2pt, mark options={mark color=cyan},  color=purple] coordinates {(1, 0.979)  (2, 0.974)  (3,0.974)  (4, 0.976)  (5, 0.978)};
			\addlegendentry{\small R@2}
			\addplot+ [smooth, mark=square*,mark size=1.2pt,mark options={mark color=cyan}, color=cyan] coordinates { (1,0.958) (2,0.950) (3,0.950) (4,0.952) (5,0.951)};
				\addlegendentry{\small MRR}
			\end{axis}
			\end{tikzpicture}
		\end{center}
    	\caption{\label{tab:Decouple} Influence of the number of Decoupling Blocks.}
\end{figure}
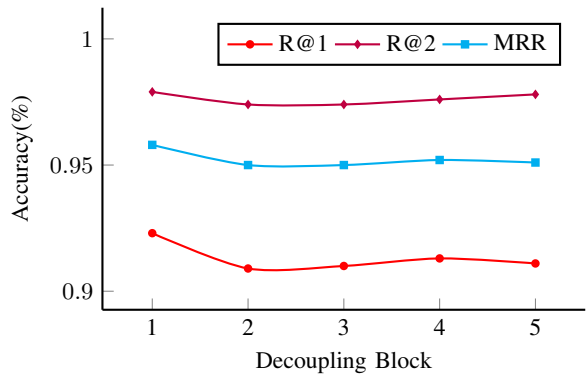

Inspired by the stacked manner of Transformer \cite{vaswani2017attention}, we have also performed a study on the effects of stacked decoupling layers. The results are shown in {{Figure \ref{tab:Decouple}}}, where we can see only one layer is enough and better than deeper. This can be explained that a deeper decoupling module is harder to learn, and only one interaction is enough.

\subsection{Number of BiGRU Layers}
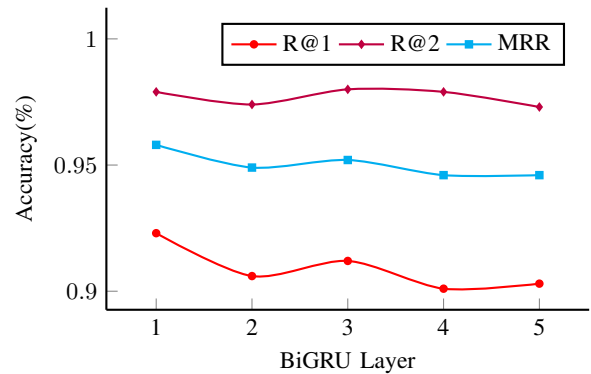
\begin{figure}
	\centering
			\setlength{\abovecaptionskip}{0pt}
			\begin{center}
			\pgfplotsset{height=5.6cm,width=6cm,compat=1.14,every axis/.append style={thick},every axis legend/.append style={ at={(0.95,0.95)}},legend columns=3 row=1} \begin{tikzpicture} \tikzset{every node}=[font=\small] \begin{axis} [width=8cm,enlargelimits=0.13, xticklabels={1,2,3,4,5}, axis y line*=left, axis x line*=left, xtick={1,2,3,4,5}, x tick label style={rotate=0},
			ylabel={Accuracy(\%)},
			ymin=0.905,ymax=1.0,
			ylabel style={align=left},xlabel={BiGRU Layer},font=\small]
			\addplot+ [smooth, mark=*,mark size=1.2pt,mark options={mark color=cyan}, color=red] coordinates
			{ (1,0.923) (2,0.906) (3,0.912) (4,0.901) (5,0.903)};
			\addlegendentry{\small R@1}
			\addplot+[smooth, mark=diamond*, mark size=1.2pt, mark options={mark color=cyan},  color=purple] coordinates {(1, 0.979)  (2, 0.974)  (3,0.980)  (4, 0.979)  (5, 0.973)};
			\addlegendentry{\small R@2}
			\addplot+ [smooth, mark=square*,mark size=1.2pt,mark options={mark color=cyan}, color=cyan] coordinates { (1,0.958) (2,0.949) (3,0.952) (4,0.946) (5,0.946)};
				\addlegendentry{\small MRR}
			\end{axis}
			\end{tikzpicture}
		\end{center}
    	\caption{\label{tab:RNN} Influence of the number of BiGRU layers on MuTual.}
\end{figure}

To see whether a deeper BiGRU will be beneficial to the modeling of a dialogue representation from the sequence of utterance representations, we have conducted experiments on the number of BiGRU layers. As shown in {{Figure \ref{tab:RNN}}}, a deeper BiGRU causes a big drop on \textbf{R}@1 metric, showing that {the shallow BiGRU is effective for capturing the information flow in the dialogue context}.

\begin{table}[t]
		\centering
        \caption{\label{tab:ablation} Ablation study on MuTual dev sets. ``UA" means utterance-aware, and ``SA" means speaker-aware.}
    {
		{
			\begin{tabular}{l c c c}
				\toprule
				\textbf{Model} & {$\textbf{R}_{4}$@1} & {$\textbf{R}_{4}$@2} & \textbf{MRR} \\
				\midrule
				ELECTRA + CDN & \textbf{0.923} & 0.979 & \textbf{0.958} \\
				\quad w/o SA-Mask & 0.909 & 0.973 & 0.949 \\
				\quad w/o UA-Mask & 0.913 & 0.977 & 0.953 \\ 
				\hdashline
				\quad w/o BiGRU (w/ Max-Pool) & 0.909 & \textbf{0.982} & 0.951 \\
				\quad w/o BiGRU (w/ Mean-Pool)  & 0.911  & 0.980 & 0.952 \\    
				\bottomrule
			\end{tabular}
		}
    }
	\end{table}

\begin{table}[t]
		\centering
		\caption{\label{tab:Gate} Influence of fusing methods on {the MuTual dev set}.}
			\setlength{\tabcolsep}{13pt}
		{
			\begin{tabular}{l c c c}
				\toprule
				\textbf{Fusing Method} & {$\textbf{R}_{4}$@1} & {$\textbf{R}_{4}$@2} & \textbf{MRR} \\
				\midrule
				ELECTRA + CDN & \textbf{0.923} & 0.979 & \textbf{0.958} \\
				\quad -Gate & 0.914 & 0.975 & 0.952 \\
				\quad -Original Info & 0.918 & 0.980 & 0.955 \\ 
				\quad \quad -Gate & 0.910 & \textbf{0.981} & 0.951 \\ 
				\bottomrule
			\end{tabular}
		}
		
	\end{table}
\subsection{Effects of Fusing Methods}
{As described in Section III-E, we use a gate to fuse the complementary information inside each channel to accumulate the information from the four decoupled channels. A simple fully-connected layer can be applied as a more simple alternative. To explore whether it is necessary to use those gates, we perform a comparative study as results are presented in Table \ref{tab:Gate}. We can see the gate mechanism yields better performance.\footnote{{It is observed that ``-Gate" yields better results towards the $\textbf{R}_{4}$@2 metric. Since the results on this metric are quite high (0.979-0.981), they are only listed for completeness. Following the existing literature \cite{whang2021response}, we focus on the more distinguishable metrics $\textbf{R}_{1}$@2 and $\textbf{MRR}$.}} In addition, we find using the information from the original sequence representation $E$ is also beneficial, which could serve as the reference for calculating the gating ratio to measure how similar is the decoupled representation with regards to the original representation.
}

\begin{table}
		\centering
		\caption{\label{tab:aggregating} Influence of the Aggregating Methods on {the MuTual dev set}. CNN uses a filter of size 3, and CNN-Multi combines filters of size 2, 3, 4, which is similar to \cite{kim2014convolutional}.}
		\setlength{\tabcolsep}{12pt}
		{
			\begin{tabular}{l c c c}
				\toprule
				\textbf{Aggregating Method} & \textbf{R@1} & \textbf{R@2} & \textbf{MRR} \\
				\midrule
				Max-Pooling & \textbf{0.923} & \textbf{0.979} & \textbf{0.958} \\
				Mean-Pooling & 0.911 & 0.975 & 0.951 \\
				CNN & 0.916 & 0.974 & 0.953 \\ 
				CNN-Multi & 0.902 & 0.974 & 0.946 \\ 
				\bottomrule
			\end{tabular}
		}
		
	\end{table}	

\subsection{Effects of Aggregating Methods}
To aggregate the word representations in one utterance, we can use simple global pooling. And another widely used method is Convolution Neural Network (CNN). Deeper CNN can capture a wide scope of the receptive field, making it successful in Computer Vision \cite{simonyan2014very} and Text Classification \cite{kim2014convolutional}. We have also compared different aggregating methods on the MuTual dev set as shown in Table \ref{tab:aggregating}. We can see that max-pooling is better than mean-pooling since it can preserve some activated signals, removing the disturbance of less important signals. However, the two CNN-based methods perform worse than max-pooling, especially those with multiple filter sizes. This can be explained that shared filters between different sentences are not flexible enough and cannot generalize well.

\subsection{Effects of Underlying Pre-trained Models}
To test the generality of the benefits of our CDN to other PrLMs, we alter the underlying PrLMs to other variants in different sizes or types. As shown in Table \ref{tab:PrLMs}, 
we see that our CDN is generally effective for the widely-used PrLMs.

\begin{table}
		\centering
		\caption{\label{tab:PrLMs} Performances of single PrLM {{baseline}} and with CDN on {the MuTual dev set}.}
\setlength{\tabcolsep}{1.3pt}
    {
		{
			\begin{tabular}{l c c c c c c}
				\toprule
				\textbf{PrLMs} & \multicolumn{2}{c}{\textbf{R@1}} & \multicolumn{2}{c}{\textbf{R@2}} & \multicolumn{2}{c}{\textbf{MRR}} \\
				 & {{Baseline}} & \textit{+CDN} & {{Baseline}} & \textit{+CDN} & {{Baseline}} & \textit{+CDN} \\
				\midrule
				{$\text{BERT}_\text{base}$} & 0.653 & \textbf{0.684} & 0.860 & \textbf{0.871} & 0.800 & \textbf{0.818} \\
				{$\text{RoBERTa}_\text{base}$} & 0.709 & \textbf{0.731} & 0.886 & \textbf{0.898} & 0.833 & \textbf{0.846} \\
				{$\text{ELECTRA}_\text{base}$} & 0.762 & \textbf{0.813} & 0.916 & \textbf{0.928} & 0.865 & \textbf{0.893} \\
			    {$\text{BERT}_\text{large}$} & 0.691 & \textbf{0.726} & 0.879 & \textbf{0.901} & 0.822 & \textbf{0.844} \\
			    {$\text{RoBERTa}_\text{large}$} & 0.834 & \textbf{0.845} & 0.952 & \textbf{0.953} & 0.908 & \textbf{0.914} \\
				{$\text{ELECTRA}_\text{large}$} & 0.906 & \textbf{0.923} & 0.977 & \textbf{0.979} & 0.949 & \textbf{0.958} \\
				\bottomrule
			\end{tabular}
		}}
		
	\end{table}
	
\begin{table}[t]
		\centering
        \caption{\label{tab:prlm-dap} Ablation study of domain-adaptive post-training on Ubuntu (ELECTRA-based CDN).}
        \setlength{\tabcolsep}{11.2pt}
    {
		{
			\begin{tabular}{l c c c}
				\toprule
				\textbf{Model} & $\textbf{R}_{10}$@1 & $\textbf{R}_{10}$@2 & $\textbf{R}_{10}$@5\\
				\midrule
				Subword  & 0.884 & 0.945 & 0.987  \\
				Span  & 0.887 & 0.946 & 0.988 \\
				WWM  & 0.889 &	0.947 & 0.989 \\
				\hdashline
				Subword + NSP & 0.912 & 0.959 & 0.992 \\
				Span + NSP & 0.912 & 0.960 & 0.992 \\
			    WWM + NSP  & \textbf{0.914} &\textbf{0.961} & \textbf{0.993} \\
				\bottomrule
			\end{tabular}
		}
    }
	\end{table}

\subsection{Influence of Domain-adaptive Post-Training}
We are interested in the influence of the post-training objectives (MLM and NSP) on overall performance. Firstly, we compare the three kinds of masking strategies, i.e., subword, span, and word level masks. Then, we evaluate the impact of the NSP objective to see if it works in the dialogue scenarios. {For the analysis, we use the Ubuntu dataset instead of the MuTual dataset because MuTual is a very small-scale dataset (only 8K context-response pairs) even with high quality, which is not appropriate for evaluating post-training. Therefore, we used Ubuntu which is widely used by existing studies (those reported in Table II), for assessing post-training.}

Table \ref{tab:prlm-dap} shows the results. We see that the contribution of the masking methods does not vary too much, and the NSP objective boosts the performance remarkably. Although NSP has been shown trivial in RoBERTa \cite{liu2019roberta} during general-purpose pre-training, it yields surprising gains in dialogue scenarios. The most plausible reason is that dialogue emphasizes the relevance between dialogue context and the subsequent response, which shares a similar goal with NSP. The findings indicate that the utterance-level language modeling would be effective for dialogue-related tasks, which would be a possible research topic in future studies.

{As pre-trained weights from the free-domain corpus are widely-used as the initialization, we are also interested in whether the pre-trained weights are dispensable when using our domain-adaptive training. The comparison between general pre-training and domain-adaptive training without the weights from the general pre-training is shown in Table \ref{tab:domain-pretrain}. We observe that training with the domain-aware corpus is much more efficient in our dialogue task. With much less training data and computation cost, our method outperforms all the baselines without large-scale pre-training on the domain-free corpus in Table \ref{table:rs_results}, and even achieves better performance than various pre-trained methods such as BERT and TADAM.}

\subsection{Application Potentials}
In terms of applications, our model can be used to build the retrieval-based chatbot, and the model latency can be tolerated. For our CDN model, the model latency mainly depends on the underlying PrLMs and batch size. Taking the MuTual dataset as an example, the inference phase is 2 min/epoch using our settings, where 886 instances are included. Such velocity is supposed to be tolerated in real-world applications. For real-world scenarios, our model can be applied to build a chatbot if responses can be first selected coarsely and then ranked by our model.

Besides the retrieval-based chatbots, we have two ways to generalize CDN to generation-based dialogue systems as (1) re-ranking module (2) the encoder in the traditional encoder-decoder framework. A generation-based model can generate some candidate responses, then to choose the better response, our model can be used to rank the candidates. Another way is using the major parts of CDN (without the final classification layer) as an encoder to learn the utterance-aware and speaker-aware representations. Then a decoder of advanced generation-based models can be utilized upon CDN to generate better responses.

\begin{table}
		\centering
        \caption{\label{tab:domain-pretrain} 
       {Comparison of of general pre-training and domain-adaptive training (without the weights from the general pre-training).} }
        \setlength{\tabcolsep}{3.6pt}
    {
		{
			\begin{tabular}{l c c c c c c c}
				\toprule
				\textbf{Model}&  \textbf{Data} & \textbf{Steps} & \textbf{Batch} & \textbf{Device} & $\textbf{R}_{10}$@1 & $\textbf{R}_{10}$@2 & $\textbf{R}_{10}$@5 \\
				\midrule
				Vanilla & 160GB & 1.75M & TPU & 2048 & 0.845 & 0.919 & 0.979 \\
				Domain & 340M & 470K & 64 & GPU & 0.821 & 0.907 & 0.979 \\
				\bottomrule
			\end{tabular}
		}
    }
	\end{table}
	
\section{Conclusion}
In this paper, we propose a novel and simple Channel-aware Decoupling Network (CDN), which decouples the utterance-aware and speaker-aware information, tackling the problem of role transition and noisy distant texts. Experiments on four retrieval-based multi-turn dialogue datasets show the superiority over existing methods. The ablation study of different sub-modules explains their effectiveness and relative importance. {We find that domain-adaptive post-training also enhances the model performance substantially. Even on the backbone with post-training, CDN still yields consistent gains, showing that CDN is generally effective no matter how strong the backbone model is.} Our work reveals a way to make better use of the semantically rich contextualized representations from pre-trained language models and gives insights on how to combine the traditional RNN models with powerful transformer-based models. In the future, we will study dialogue-aware pre-training techniques and in-depth modeling of dialogue structures such as discourse-aware parsing and abstract meaning representation for dialogue texts.

\ifCLASSOPTIONcaptionsoff
  \newpage
\fi

\bibliographystyle{IEEEtranN}
\bibliography{mdfn}

	\begin{IEEEbiography}[{\includegraphics[width=1in,height=1.25in,clip,keepaspectratio]{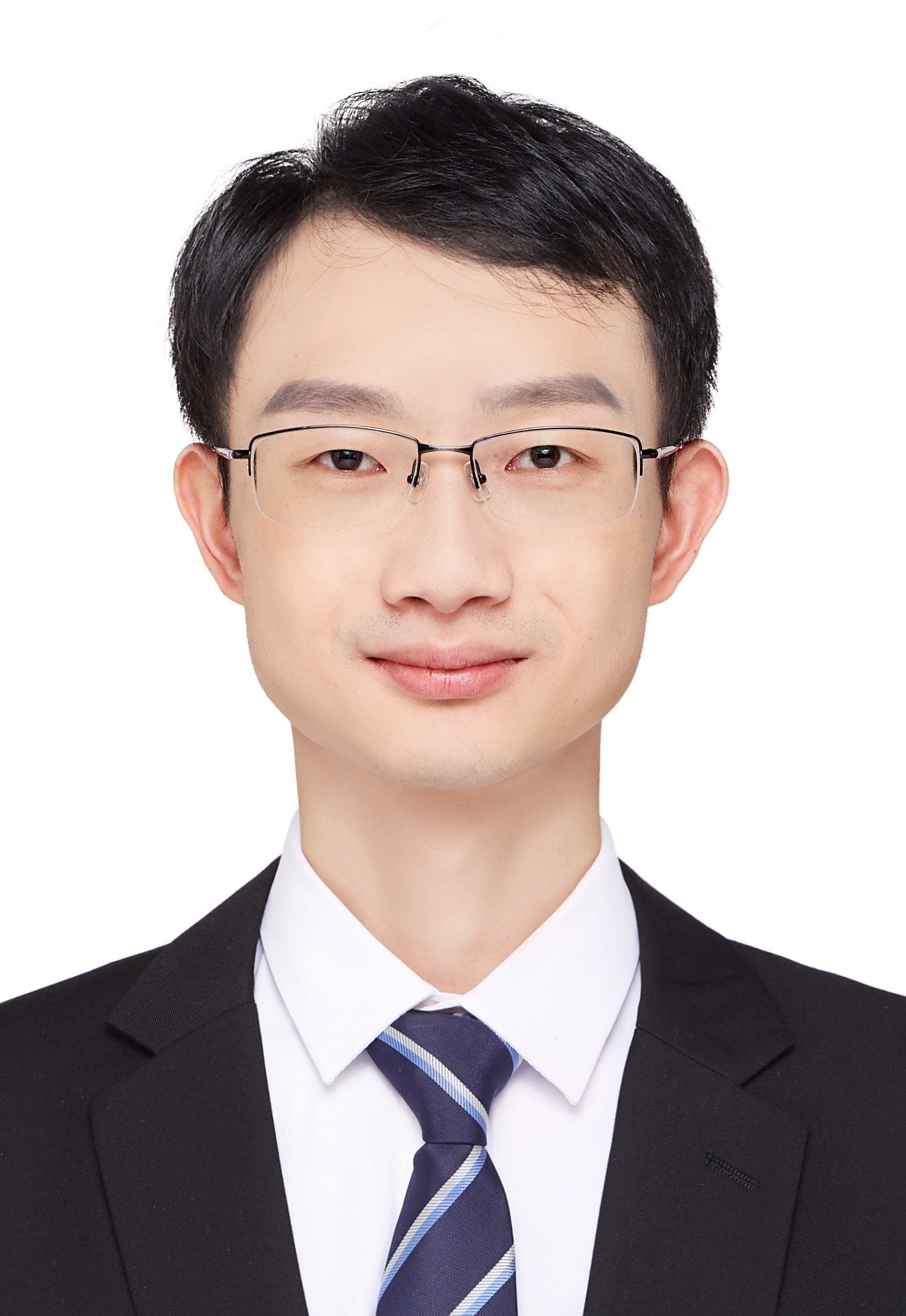}}]{Zhuosheng Zhang}
  		received his Bachelor's degree in internet of things from Wuhan University in 2016, his M.S. degree in computer science from Shanghai Jiao Tong University in 2020. He is working towards the Ph.D. degree in computer science with the Center for Brain-like Computing and Machine Intelligence of Shanghai Jiao Tong University. He was an internship research fellow at NICT from 2019-2020. His research interests include natural language processing, machine reading comprehension, and dialogue systems. 
	\end{IEEEbiography}
    
	\begin{IEEEbiography}[{\includegraphics[width=1in,height=1.25in,clip,keepaspectratio]{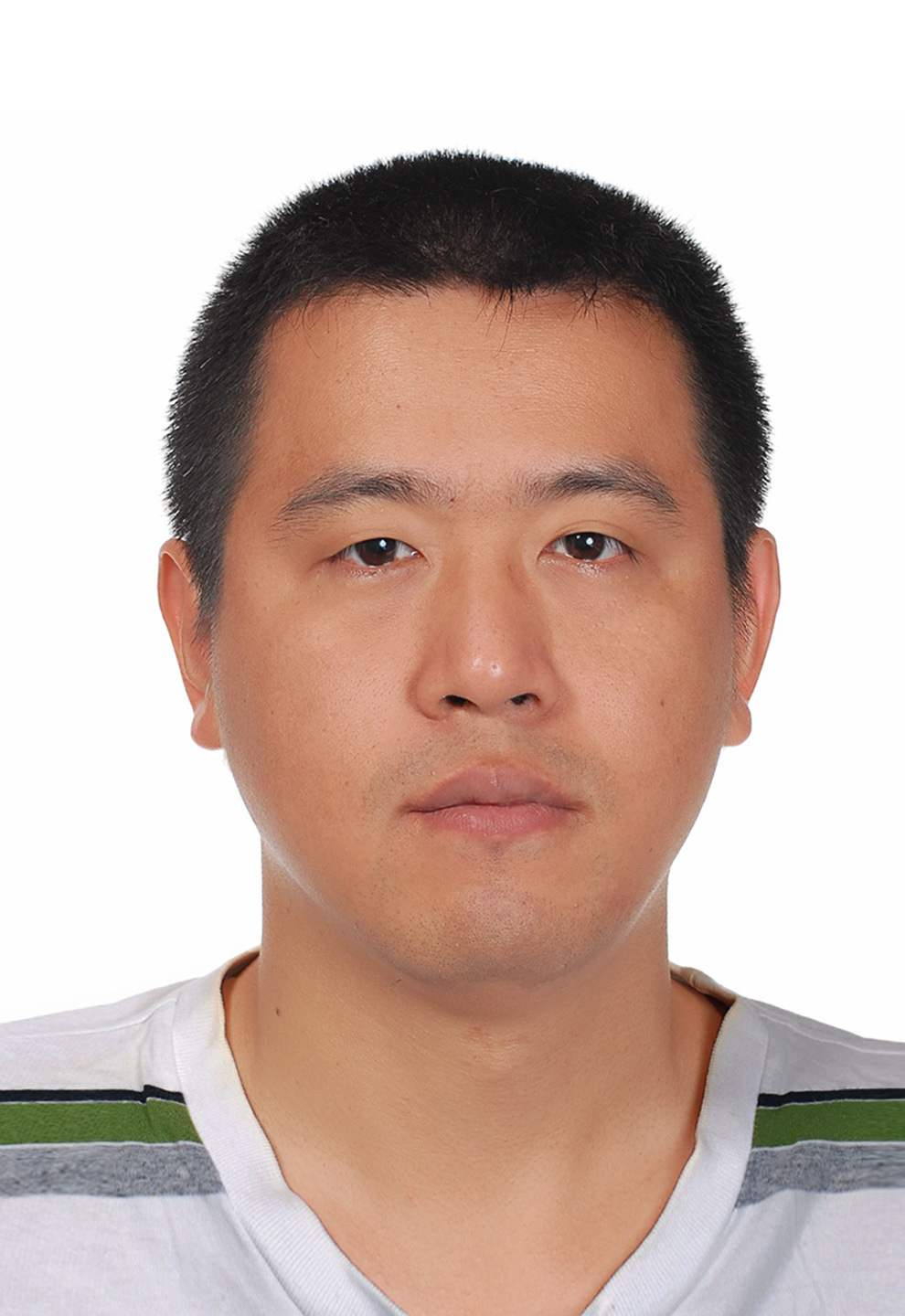}}]{Hai Zhao}
		received the BEng degree in sensor and instrument engineering, and the MPhil degree in control theory and engineering from Yanshan University in 1999 and 2000, respectively,
		and the PhD degree in computer science from Shanghai Jiao Tong University, China in 2005. 
		He is currently a full professor at department of computer science and engineering,  Shanghai Jiao Tong University after he joined the university in 2009. 
		He was a research fellow at the City University of Hong Kong from 2006 to 2009, a visiting scholar in Microsoft Research Asia in 2011, a visiting expert in NICT, Japan in 2012.
		He is an ACM professional member, and served as area co-chair in ACL 2017 on Tagging, Chunking, Syntax and Parsing, (senior) area chairs in ACL 2018, 2019 on Phonology, Morphology and Word Segmentation.
		His research interests include natural language processing and related machine learning, data mining and artificial intelligence.
	\end{IEEEbiography}
	\begin{IEEEbiography}[{\includegraphics[width=1in,height=1.25in,clip,keepaspectratio]{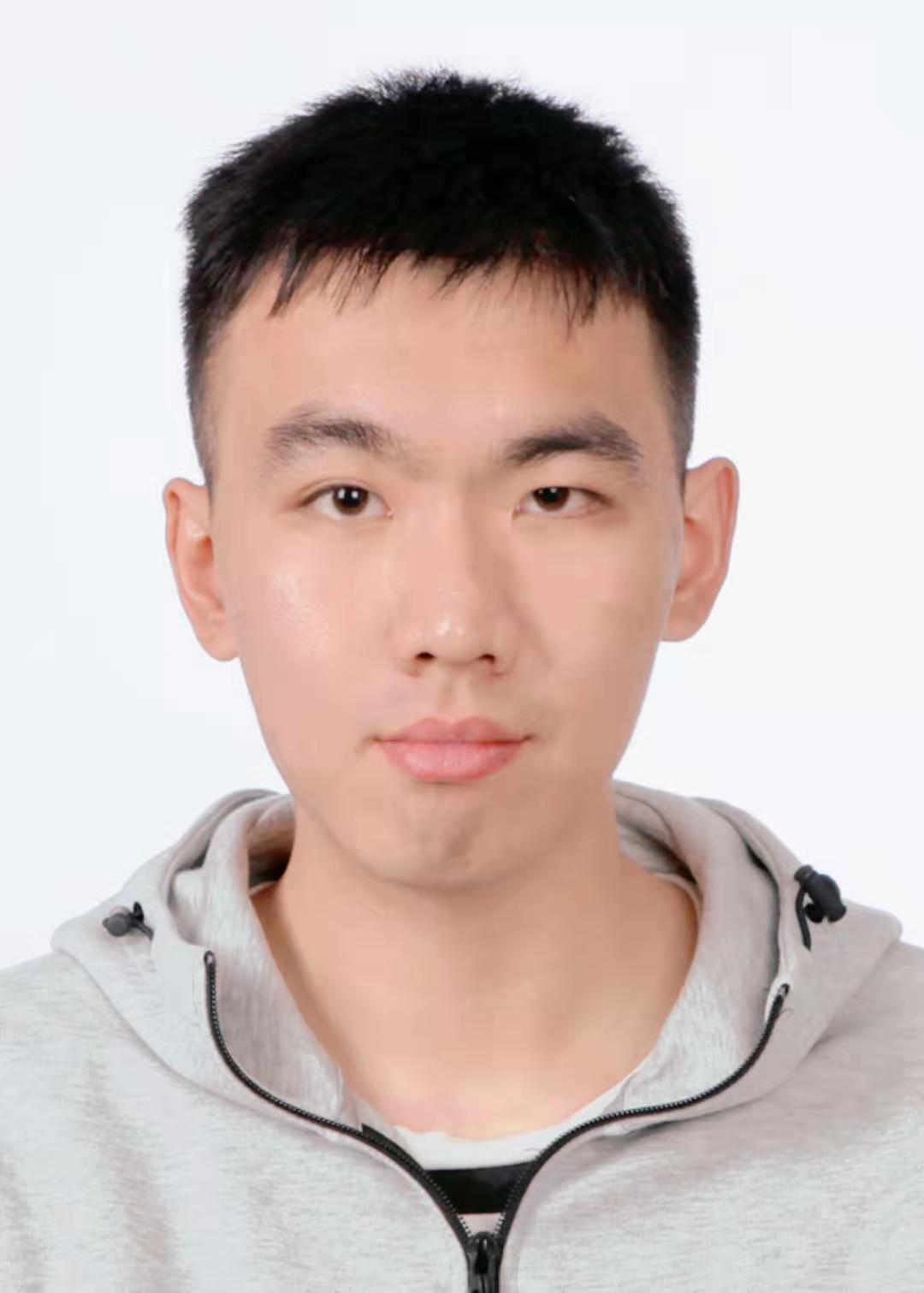}}]{Longxiang Liu}
	    received Bachelor's degree in the department of computer science and engineering from Shanghai Jiao Tong University in 2021. He is now a graduate student studying in the University of Chinese Academy of Sciences. He is working towards the M.S. degree in computer science with the Natural Language Processing Group of Key Laboratory of Intelligent Information Processing of Institute of Computing Technology in Chinese Academy of Sciences. He has done researches in Center for Brain-like Computing and Machine Intelligence of Shanghai Jiao Tong University. His research interests include natural language processing, machine reading comprehension and dialogue systems.
	\end{IEEEbiography}

\end{document}